\pdfoutput=1

\documentclass[11pt]{article}

\usepackage[]{acl}

\usepackage{times}
\usepackage{latexsym}

\usepackage{booktabs}
\usepackage{graphicx}
\usepackage{subfigure}
\usepackage{todonotes}
\usepackage{multirow}
\usepackage{hyperref}
\usepackage{amsmath}
\usepackage{comment}

\iffalse
\newcommand{\wojciech}[1]{}
\newcommand{\af}[1]{}
\newcommand{\jw}[1]{}
\newcommand{\arti}[1]{}
\else
\newcommand{\wojciech}[1]{\textcolor{orange}{\textbf{Wojciech:} #1}}
\newcommand{\af}[1]{\textcolor{blue}{\textbf{Alex:} #1}}
\newcommand{\jw}[1]{\textcolor{green}{\textbf{Jason:} #1}}
\newcommand{\arti}[1]{\textcolor{red}{\textbf{Arti:} #1}}
\fi

\newcommand{\squality}{SQuALITY}
\newcommand{\socratic}{\textsc{Socratic}}
\newcommand{\askanswer}{\emph{Ask\&Answer}}
\newcommand{\ask}{\emph{Ask}}
\newcommand{\answer}{\emph{Answer}}

\usepackage[T1]{fontenc}

\usepackage[utf8]{inputenc}

\usepackage{microtype}

\title{\socratic{} Pretraining: Question-Driven Pretraining \\ for Controllable Summarization}

\author{
 Artidoro Pagnoni\Thanks{~~Work done during internship at Salesforce} $^{1}$ 
 \quad \textbf{Alexander R. Fabbri}$^{2}$
 \quad \textbf{Wojciech Kry\'sci\'nski}$^{2}$ 
 \quad \textbf{Chien-Sheng Wu}$^{2}$ \\
  $^1$University of Washington, 
  $^2$Salesforce AI Research\\
  \text{artidoro@uw.edu}, \text{\{afabbri, wojciech.kryscinski, wu.jason\}@salesforce.com} 
 }

\begin{document}
\maketitle

\begin{abstract}
In long document controllable summarization, where labeled data is scarce, pretrained models struggle to adapt to the task and effectively respond to user queries.
In this paper, we introduce \socratic{} pretraining, a question-driven, unsupervised pretraining objective specifically designed to improve controllability in summarization tasks. 
By training a model to generate and answer relevant questions in a given context, \socratic{} pretraining enables the model to more effectively adhere to user-provided queries and identify relevant content to be summarized.
We demonstrate the effectiveness of this approach through extensive experimentation on two summarization domains, short stories and dialogue, and multiple control strategies: keywords, questions, and factoid QA pairs.
Our pretraining method relies only on unlabeled documents and a question generation system and outperforms pre-finetuning approaches that use additional supervised data.
Furthermore, our results show that \socratic{} pretraining cuts task-specific labeled data requirements in half, is more faithful to user-provided queries, and achieves state-of-the-art performance on QMSum and \squality{}.

\end{abstract}

\section{Introduction}

\begin{figure}[t]
    \centering
    \includegraphics[width=\columnwidth]{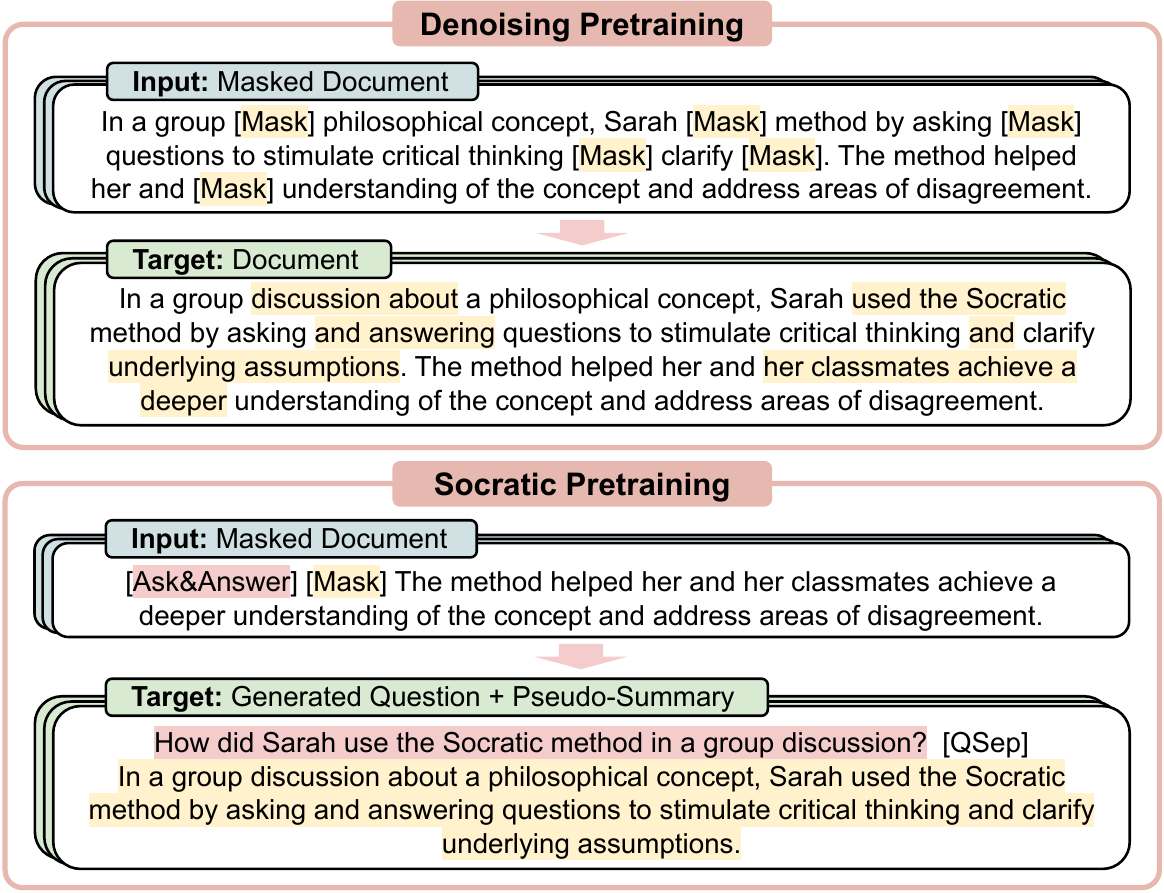}
    \caption{Our \socratic{} pretraining compared to denoising. We mask important sentences in unlabeled input documents and train the model to generate both \textbf{questions} and \textbf{pseudo-summaries} as their answers. 
    }
    \label{fig:socratic_pretraining}
\end{figure}
Summarization systems are designed to help users navigate large amounts of information \citep{edmunds2000problem}, but often fail to meet the unique needs of different users, especially for long documents. Recent research has explored ways to make summarization systems more controllable \citep{bornstein1999interactive, leuski-etal-2003-ineats} by allowing users to input queries or control sequences such as keywords \citep{he2020ctrlsum}, questions \citep{zhong-etal-2021-qmsum}, entity chains \citep{narayan-etal-2021-planning}, or question-answer pairs \citep{narayan2022conditional}.

A challenge shared by all of the mentioned approaches is the absence of abundant labeled data. Currently available datasets for training these systems are the result of expensive annotation efforts \citep{zhong-etal-2021-qmsum, kulkarni2020aquamuse, wang2022squality} with only hundreds to a few thousand query-document pairs, with the same document often being repeated. 
This translates into poor adherence of generated summaries to user-provided queries, particularly when these are finegrained plans.
Recent work demonstrates the benefits of tailoring the pretraining objective to downstream task characteristics, especially where training data is difficult to obtain in large quantities like factuality-focused and multi-document summarization \citep{wan-bansal-2022-factpegasus, xiao-etal-2022-primera}. 
In controllable summarization, summaries are grounded by queries, so designing an objective for the task requires introducing realistic queries in unlabeled data in a scalable manner.

This work introduces \socratic{} pretraining, an unsupervised pretraining objective for language models that is specifically designed for controllable summarization. It is inspired by the Socratic method and aims to facilitate the identification of relevant content and ensure that the generated summary faithfully responds to the user query.
During \socratic{} pretraining (see \autoref{fig:socratic_pretraining}) the language model is trained to generate relevant questions based on an input document and then answer them, bringing finegrained controllability to model pretraining which translates to better adherence to user queries. 

\socratic{} pretraining only relies on unlabeled data and a question generation system and outperforms pre-finetuning approaches relying on additional supervised data \citep{aghajanyan-etal-2021-muppet, wei2021finetuned, fabbri-etal-2021-improving}.
In this work, we demonstrate the effectiveness of the \socratic{} objective through \emph{pretraining adaptation}, where a language model is further pretrained with the \socratic{} objective before finetuning on task-specific labeled data.

In summary, our contributions are as follows\footnote{Our code is be available 
at \url{https://github.com/salesforce/socratic-pretraining}
}:
\begin{itemize}
    \item We introduce the \socratic{} pretraining objective for controllable summarization to improve adherence to user-specified queries or plans, both high-level and finegrained.
    \item We show that \socratic{} pretraining performs well across domains, control strategies, and achieves state-of-the-art performance on two datasets.
    \item We perform ablations on our approach showing that \socratic{} pretraining cuts labeled data requirements in half.
\end{itemize}

\section{Related Work}
\paragraph{Task-Specific Pretraining Adaptation} 
Current state-of-the-art methods in abstractive summarization apply a two-step approach where models are first pretrained on large corpora of text with task-agnostic variations of the text denoising objective and next finetuned on labeled examples from the target task \citep{lewis-etal-2020-bart, 2020t5}. 
\par 
However, in tasks where labeled data is scarce, task-specific pretraining objectives have been shown to provide significant benefits. Recent work adapted language models to summarize multiple documents \citep{xiao-etal-2022-primera}, produce more factual summaries \citep{wan-bansal-2022-factpegasus}, or plan with entity chains \citep{narayan-etal-2021-planning}. 
We build on these methods, focusing on the downstream task of controllable summarization.
\par
Other studies demonstrate the effect of continued pretraining \citep{gururangan-etal-2020-dont} and pre-finetuning \citep{aghajanyan-etal-2021-muppet, wei2021finetuned, fabbri-etal-2021-improving} on downstream task adaptation. These either continue training with the same objective on data in the downstream task domain or perform multitask learning using labeled data.
In this work, we demonstrate the benefits of language model adaptation with a task-specific pretraining objective without additional supervised data and show that these benefits are consistent and statistically significant in low-resource settings like query-focused summarization (QFS).

\paragraph{Controllable Summarization}
Controllable text generation \citep{hu2017toward} aims to control properties of the generated text including style \citep{kumar2021controlled}, length, or content \citep{fan-etal-2018-controllable, he2020ctrlsum}. Approaches for content control vary according to the type of control: keywords \citep{he2020ctrlsum}, entities \citep{narayan-etal-2021-planning}, questions \citep{vig-etal-2022-exploring}, factoid question-answer pairs (also called QA blueprints) \citep{narayan2022conditional}. As opposed to methods like GSum \citep{dou-etal-2021-gsum}, which insert control tokens on the encoder side, we focus on decoder-based methods which do not require re-encoding the document when the control sequences are updated. In summarization, these controls can broadly indicate the information to summarize, like the questions in query-focused summarization, or provide a detailed plan of the text to be generated, like the entity chains. While these types of control are not typically studied together we show that our \socratic{} pretraining provides benefits across the board for both high-level and finegrained queries and plans.  

\paragraph{Learning with Questions}
Inspired by the Socratic method, recent literature in education theory shows students generate questions as a way of learning \citep{teaching-students-to-generate-questions, students-generating-questions-as-a-way-of-learning}, hinting at the potential benefits that could derive from incorporating questions during model training. 
Previous work shows that question-answer pairs, both generated \citep{du-etal-2017-learning, alberti-etal-2019-synthetic, ko2021discourse, murakhovska-etal-2022-mixqg, chakrabarty2022consistent} and from the web \citep{narayan2020qurious}, can provide useful training signal for pretrained encoders \citep{jia-etal-2022-question} as well as question generation and abstractive summarization systems \citep{narayan2022conditional}.
Our \socratic{} objective builds on these observations and is designed to improve sequence-to-sequence model pretraining for more controllable summarization systems. Similar to information-seeking Dialogue Inpainting \citep{dai2022dialog}, \socratic{} pretraining extracts questions from unlabeled data focusing on higher-level questions, whose answers are full sentences, instead of factoid QA pairs.

\section{\socratic{} Pretraining}
\label{sec:socratic-pretraining}
During \socratic{} pretraining, the model takes as input a document with important sentences masked and is trained to generate questions about the masked content and produce the mask itself.
As seen in \autoref{fig:socratic_pretraining}, \socratic{} pretraining is formulated as a sequence-to-sequence task and consists of two steps 1) important content is selected from unlabeled documents to be masked, and 2) a question-generation system is applied to produce questions about the selected content.
The question augmentation component trains the model to produce summaries grounded to questions and allows for controllability as the end-user can prompt the model decoder with new questions during inference. 
We describe both steps below.

\subsection{Content Selection}
Selecting important content is essential for the model to learn to generate salient questions and summaries. In \socratic{} pretraining, this content selection is done using the PEGASUS-style Gap Sentence Generation (GSG) objective \citep{pegasus}, which we now briefly describe. 
Sentences with the highest self-Rouge with the document are selected for masking, ensuring that there is high information overlap with the rest of the document. 
The selected sentences, concatenated, produce a \emph{pseudo-summary} of the document. 
As in PEGASUS, a Gap Sentence Ratio (GSR) of 45\% is used, meaning that 45\% of the sentences in the document are selected to appear in the target pseudo-summary. To help the model learn to copy, 80\% of these sentences are masked and 20\% are kept unmasked in the input document. Documents and summaries are truncated to 512 and 256 tokens.

\subsection{Question Augmentation}
After selecting the pseudo-summary, a question generation (QG) system is applied to obtain a question from each sentence of the pseudo-summary. The QG system takes as input one of the selected sentences at a time and the unmasked document as context.
We apply MixQG \citep{murakhovska-etal-2022-mixqg}, a state-of-the-art QG system.

The choice to generate a question for each selected sentence, as opposed to each entity or the entire summary, is driven by three reasons.
First, sentences in the pseudo-summary are selected from across the document and generally lack coherence, so there is no single query they collectively answer.
Second, current QG systems are not trained to produce paragraph-level questions.
Third, entity-level questions are often simple paraphrases of the answer sentence and are uncommon in QFS datasets.

Questions whose answers are full sentences, therefore, offer a compromise in terms of the complexity of the question and the coherence of the answer. We refer to these sentence-level questions as \emph{content-questions} as they tend to ask about the content of the document instead of specific entities.
\begin{figure}[t]
    \centering
    \includegraphics[width=0.9\columnwidth]{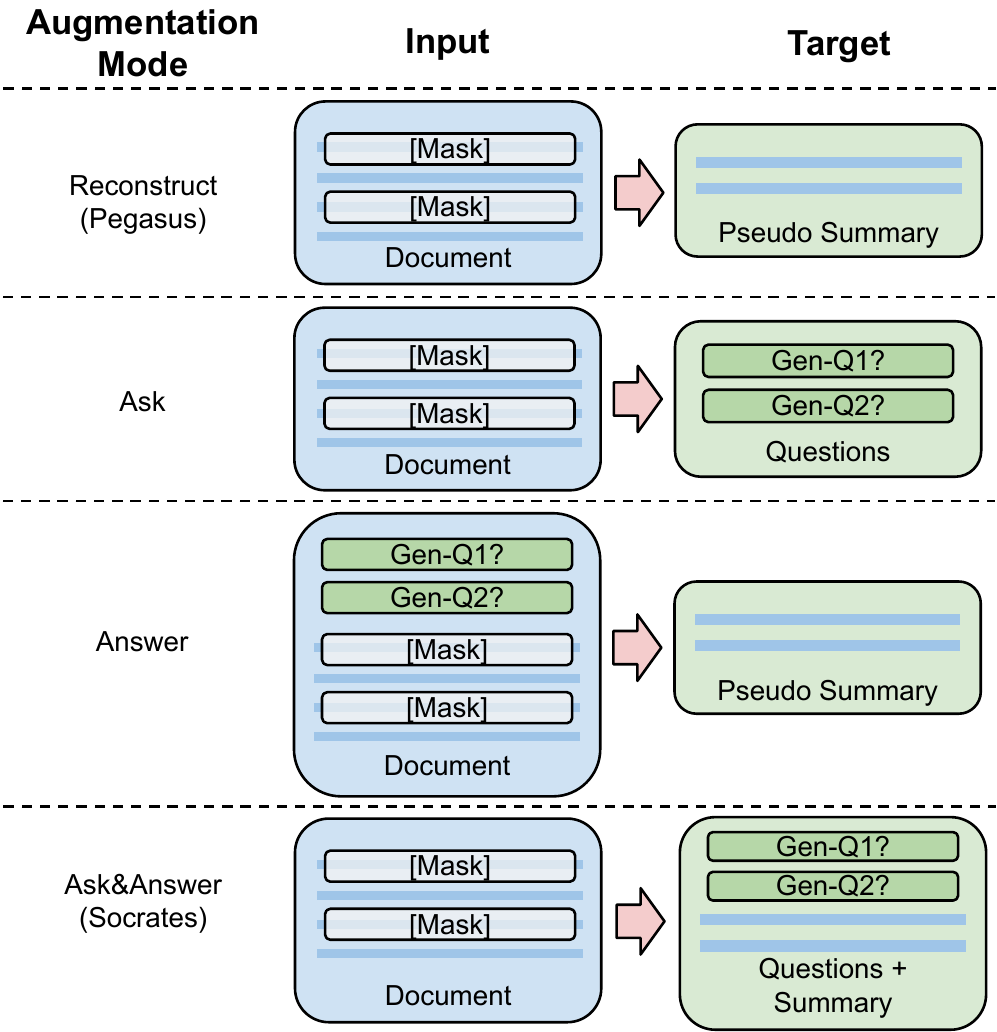}
    \caption{\socratic{} augmentation modes vs. Pegasus.}
    \label{fig:augmentation_mode}
\end{figure}
\subsection{Training Objective}
After obtaining the questions, there are multiple ways to introduce them in the training objective either in the input or in the target text. As seen in \autoref{fig:augmentation_mode}, we experiment with three modes on top of the base GSG objective:
\begin{itemize}
    \item \emph{Reconstruct.} The reconstruct mode is the default GSG mode where no questions are introduced. The masked document is the input and the pseudo-summary is the target text. We provide this mode as a baseline for our approach.
    \item \emph{Ask.} Given the masked document as input, the model is trained to only predict the questions about the masked sentences. This is the only mode where the target text does not include the pseudo-summary. With this mode, the model is trained to predict which questions can be asked in a given context. 
    \item \emph{Answer.} Here, the questions are prepended to the masked input document while the target text remains the pseudo-summary. 
    This mode is similar to how queries are introduced to the model during query-focused summarization and should help the model learn to respond to user-provided queries. 
    However, this mode forgoes content planning as each generated sentence corresponds to one of the questions prepended to the input. 
    \item \emph{Ask\&Answer.} This mode combines benefits from both \ask{} and \answer{} modes. The model is tasked to first generate questions about the document and then, conditioning on both the document and the generated questions, the pseudo-summary. The model conditions on the generated questions in the decoder. This mode can be seen as first generating a finegrained plan for the pseudo-summary and then the pseudo-summary itself.
\end{itemize}

Like \citet{tay2022unifying}, we prepend special tokens \texttt{<ask>}, \texttt{<answer>}, and \texttt{<ask\&answer>} to the input document to specify the augmentation mode, and the \texttt{<qsep>} token to separate the generated questions from the target pseudo-summary.

\section{Experimental Setup}
\label{sec:experiments}
We describe the experimental setup that we use to study \socratic{} pretraining along with empirical studies justifying our design decisions. 

\subsection{Model Architecture}
The \socratic{} objective can be applied to any sequence-to-sequence language model irrespective of its specific architecture. In our experiments, we choose BART-large \citep{lewis-etal-2020-bart}, as the starting point for \socratic{} pretraining adaptation. 
Following previous work on pretraining adaption for summarization, we pick BART over PEGASUS for its smaller size without performance compromises on summarization benchmarks and its more general-purpose pretraining objective.
BART is also the underlying model in the SegEnc \citep{vig-etal-2022-exploring} architecture, which achieved state-of-the-art performance on QMSum, outperforming models such as LongT5 \citep{guo-etal-2022-longt5}.

Instead of pretraining the language model from scratch, we demonstrate the effectiveness of the proposed objective through what we call \emph{pretraining adaptation}, where a generic language model is further pretrained with the \socratic{} objective before being finetuned on task-specific labeled data. 
Although we introduce a new term for this training phase, \emph{pretraining adaptation} was recently employed to evaluate task-specific pretraining objectives for factuality and multi-document summarization \citep{wan-bansal-2022-factpegasus, xiao-etal-2022-primera}. 

After \socratic{} pretraining adaptation, the resulting model is used to initialize the SegEnc architecture, which is then finetuned on labeled data from downstream tasks. Pretraining and finetuning hyperparameter details are available in \ref{sec:training-details}.
\subsection{Pretraining Corpus}
We experiment with three different corpora, two of which are part of the Pile \citep{gao2020pile}. 
\begin{itemize}
    \item \emph{OpenWebText2} is a web-scraped dataset inspired by WebText \citep{radford2019language} that uses Reddit upvotes of outgoing links as a proxy for page quality. 
    \citet{2020t5} found this dataset to work well for summarization pretraining.
    \item \emph{Books3} is a collection of both fiction and non-fiction books. 
    We explore this data because our downstream tasks involve the short story and dialogue domains, and \citet{csaky-recski-2021-gutenberg} show books can be a good source of dialogue data.
    \item \emph{UnDial} \citep{he2022galaxy} We also explore using a dialogue corpus.
    As there are only two speakers in each dialogue in UnDial, we use a simple rule-based system to convert dialogues to third person.
    The pseudo-summary and related questions are then expressed in the third person while the input remains in the original dialogue format. 
\end{itemize}

\subsection{Downstream Tasks}
To determine whether \socratic{} pretraining improves model initialization for finetuning on controllable summarization, we test on two downstream datasets for query-focused, long-document summarization: QMSum and \squality{} (dataset statistics can be found in \ref{sec:dataset-info}). We focus on long document datasets as a challenging and practical testbed for controllable summarization methods. 

\paragraph{QMSum.}
QMSum is a benchmark for query-based, multi-domain meeting summarization \citep{zhong-etal-2021-qmsum}. The dataset consists of 1,808 query-summary pairs over 232 meetings, including product, academic, and parliamentary meetings.
\paragraph{\squality{}.}
\squality{} is a dataset for query-based short stories summarization \citep{wang2022squality}. The dataset is composed of 625 examples over 100 stories with four long reference summaries per document-question pair. 

\subsection{Evaluation Protocol}
We apply the standard Rouge \citep{lin-2004-Rouge} and BERTScore \citep{bert-score} metrics to compare model generations with reference summaries on downstream finetuning tasks. In \squality{}, we use the same procedure as the dataset authors to incorporate multiple references by taking the maximum score over the reference summaries. 
We also conduct a human evaluation study to ensure the variations between models are meaningful to users. Details on the setup can be found in \ref{sec:human-evaluation}.
\begin{figure}
    \centering
    \includegraphics[width=0.8\columnwidth]{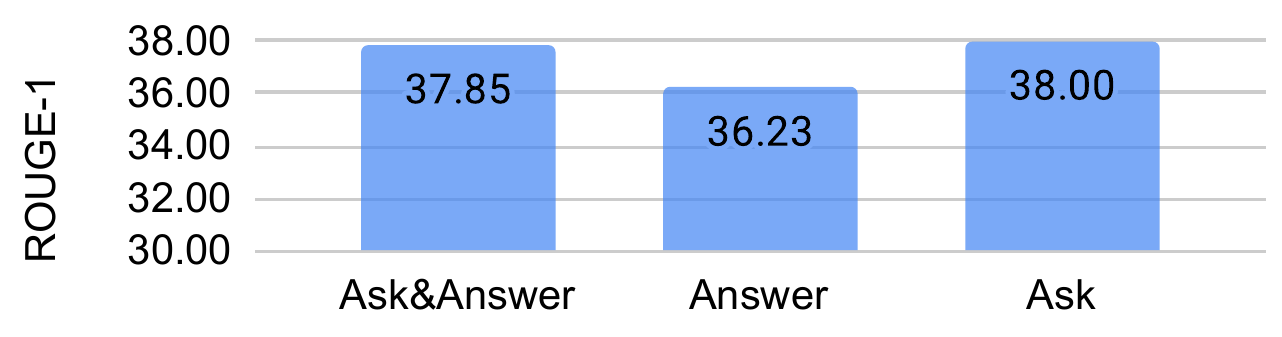}
    \caption{Comparison of question augmentation modes.
    }
    \label{fig:augmentation_mode_results}
\end{figure}
\begin{figure}
    \centering
    \includegraphics[width=0.9\columnwidth]{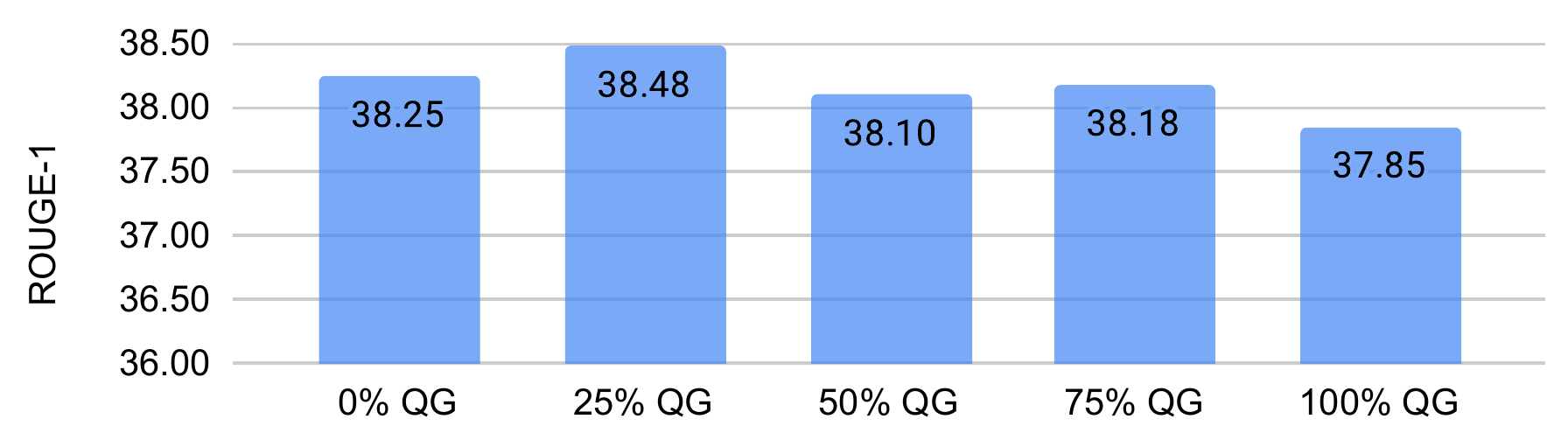}
    \caption{Comparison of QG augmentation proportions.
    }
\label{fig:augmentation_proportion}
\end{figure}
\section{\socratic{} Pretraining Ablations}
In this section, we corroborate our design choices with ablation studies of the components of \socratic{} pretraining. 
Similar to \citet{pegasus} and \citet{2020t5}, to save time and resources, we conduct the ablations of the objective on a small scale by restricting the pretraining adaptation to 1M documents from the OpenWebText2 corpus and then finetuning it on the full downstream task datasets. We report the mean over five randomly initialized finetuning runs on the validation set.

\paragraph{Question Augmentation Modes} 
\label{sec:augmentation_comparison}
In \autoref{fig:augmentation_mode_results}, we compare the performance of the three approaches for incorporating generated questions in the \socratic{} objective.
The \ask{} and \emph{Ask\&Answer} perform similarly while \answer{} lags behind. 
This is in line with our hypothesis that learning which questions are relevant in a given context is a useful training signal for the model.
The \emph{Ask\&Answer} mode also grounds the pseudo-summary generation in a sequence of finegrained questions. Therefore, it is chosen to be used in \socratic{} pretraining.

\begin{figure}[t!]
\subfigure[QMSum]{\includegraphics[width=0.48\columnwidth,trim={0 2cm 0 0},clip]{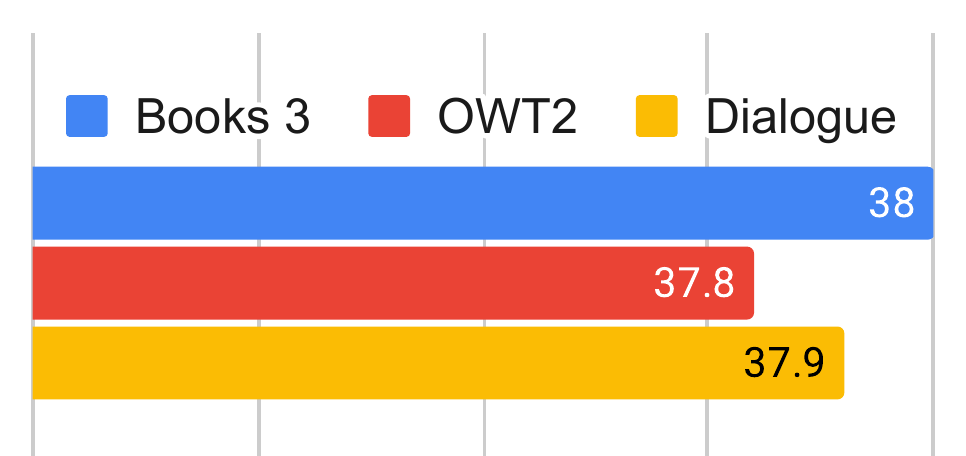}}
\subfigure[Squality]{\includegraphics[width=0.48\columnwidth,trim={0 2cm 0 0},clip]{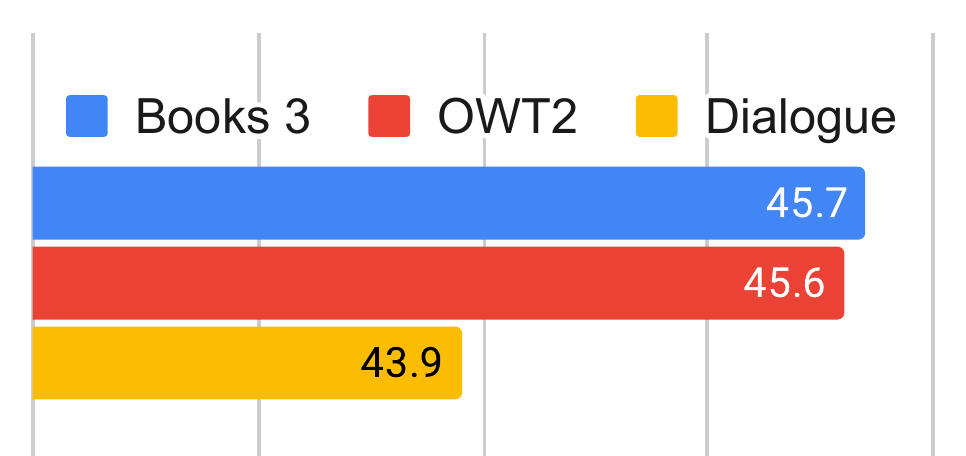}}
\vspace{-0.9\baselineskip}
\caption{Effect of the pretraining corpus (dev set).}
\label{fig:domains}
\end{figure}

\paragraph{Question Augmentation Proportion} 
Incorporating questions with the \emph{Ask\&Answer} mode in each pretraining example could bias the model to always start by generating questions. 
We hypothesize that combining the \emph{Reconstruct} mode with the \emph{Ask\&Answer} mode could alleviate this bias.  
In \autoref{fig:augmentation_proportion}, we find that introducing questions in 25\% of pretraining examples leads to the best performance and use this proportion when scaling the pretraining adaptation.

\paragraph{Pretraining Corpus Selection}
In \autoref{fig:domains}, we find that the choice of pretraining corpus has a small but consistent effect on the performance of the \socratic{} pretrained model on downstream tasks. The Books3 corpus performs best both on QMSum and \squality{}. The dialogue corpus offers a slight advantage over OpenWebText2 on QMSum, a dialogue summarization task, while the opposite is true for \squality{}. 
As a result, the full Books3 corpus, consisting of 30M training instances, is used in further experiments.

\section{Query Focused Summarization Results}
\label{sec:results}
We scale the \socratic{} pretraining adaptation based on the findings of the previous ablation and evaluate its downstream effects on query-focused summarization.
Unless specified, the results in this section are averaged over five randomly initialized finetuning runs on the downstream tasks.

In \autoref{tab:orig_results}, we compare the effect of \socratic{} pretraining to other pretraining strategies on  QMSum and \squality{}. We obtain an improvement of +1.01 and +0.53 Rouge-1, respectively, surpassing even the use of additional supervision from the related dataset WikiSum in \citet{vig-etal-2022-exploring} and achieving new state-of-the-art results. These improvements are validated by a human study reported in \autoref{fig:humaneval_qfs} and showing that \socratic{} SegEnc performs better than the baselines in 59-65\% of instances. Details of the human evaluation are found in \ref{sec:human-evaluation}.

\begin{table}[t!]
    \centering
    \resizebox{\columnwidth}{!}{
    \begin{tabular}{lrrrr}
    \toprule
    Model & Rouge1 & Rouge2 & RougeL & BS-R\\
    \midrule
    \textbf{QMSum} \\
    \midrule
    BART-LS \citep{xiong2022adapting} & 37.90 & 12.10 & 33.10 & - \\
    \midrule
    BART-Large SegEnc  & 37.05 & 13.04 & 32.62 & 87.44 \\
     \textit{+ WikiSum Pre-Finetuning} & \textit{37.80} & \textit{13.43} & \textit{33.38} & - \\
    \midrule
    + BART Pret. 1M & 36.64 & 12.44 & 31.94 & 86.94  \\
    + \socratic{} Pret. 1M & 37.46 & 13.32 & 32.79 & 87.54 \\
    \midrule
    + PEGASUS Pret. & 37.29 & 13.30 & 32.70 & 87.48  \\
    + \socratic{} Pret. & \textbf{38.06} & \textbf{13.74} & \textbf{33.51} & \textbf{87.63}  \\
    \midrule
    \textbf{Squality} \\
    \midrule
    LED & 27.7 & 5.9 & 17.7 & -  \\
    PEGASUS & 38.2 & 9.0 & 20.2 & -  \\
    BART & 40.2 & 10.4 & 20.8 & -  \\
    BART + DPR & 41.5 & 11.4 & 21.0 & - \\
    Human & 46.6 & 12.5 & 22.7 & - \\
    \midrule
    BART-Large SegEnc & 45.68 & 14.51 & 22.47 & 85.86 \\
    + PEGASUS Pret. & 45.78 & 14.43 & \textbf{22.90} & 85.94  \\
    + \socratic{} Pret. & \textbf{46.31} & \textbf{14.80} & 22.76 & \textbf{86.0}4 \\
    \bottomrule
    \end{tabular}
    }
    \caption{Results on QMSum and \squality{} with pretraining on Books3. Baselines from  \citet{vig-etal-2022-exploring} and \citet{wang2022squality} respectively. 
    1M indicates that 1M pretraining instances are used.
    }
    \label{tab:orig_results}
\end{table}

\subsection{Disentangling the Effect of Questions}

The main baseline for \socratic{} pretraining is the PEGASUS style GSG pretraining. 
We therefore perform a pretraining adaptation of BART-large with the GSG objective on the full Books3 corpus. 
In \autoref{tab:orig_results}, we observe that GSG pretraining on the full Books3 corpus improves by +0.24 Rouge-1 over the BART SegEnc model.
However, with the \socratic{} objective, 1M examples from Books3 (1/30 of the full corpus) are sufficient to surpass GSG pretraining, with a +0.41 Rouge-1 improvement over BART SegEnc. 
This indicates that GSG pretraining, tailored to generic summarization, is only marginally helpful in tasks where summaries have to answer user-provided queries. In addition, increasing the corpus for \socratic{} pretraining to the entire Books3 corpus further improves the performance by +0.60 Rouge-1 on QMSum, showing that the benefits of the pretraining objective do not saturate early and that the model continues to improve with additional \socratic{} pretraining. 

We also compare to BART-LS, an orthogonal approach that tailors BART's architecture, pretraining corpus, and objective to long documents \citep{xiong2022adapting}.  
While our approaches are complementary, we outperform BART-LS on QMSum by +1.64 Rouge-2.  
This confirms our hypothesis that grounding generations in control queries in \socratic{} pretraining is beneficial in controllable summarization, even more so than better long document modeling. 

\subsection{Comparing to Continued Pretraining}
\citet{gururangan-etal-2020-dont} show that language models can be successfully adapted to the task domain by continuing to pretrain them in the new domain.
This raises the question of whether improvements due to \socratic{} pretraining are simply due to a better affinity of the pretraining corpus to the task domain. 
To answer this question, we perform continued pretraining\footnote{For consistency, we use \href{https://github.com/facebookresearch/fairseq}{Fairseq} to pretrain BART-large} on a 1M subset of the Books3 corpus and next finetune the model on QMSum. \autoref{tab:orig_results} shows that continued pretraining slightly hurts Rouge-1 performance. In comparison, performing \socratic{} pretraining on the same corpus improves performance by +0.41 Rouge-1.
This observation rules out that improvements achieved through \socratic{} pretraining are simply due to improved domain adaptation.


\begin{figure}[t!]
    \centering
    \includegraphics[width=1.0\columnwidth, trim= {0 0.8cm 0 0.3cm}, clip]{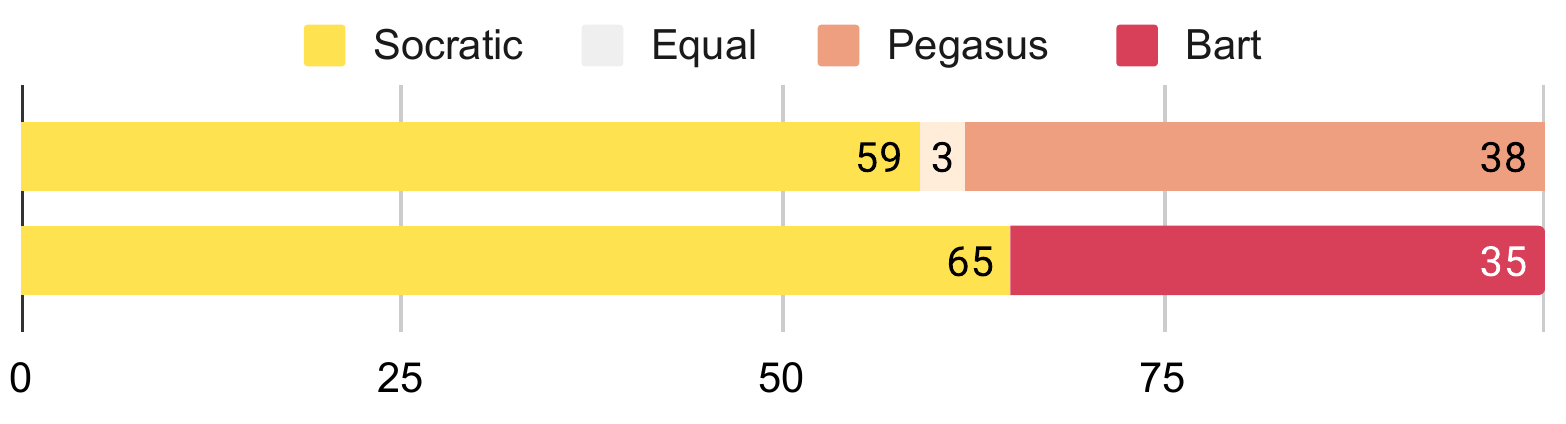}
    \caption{Human annotators' preferences on QMSum.}
    \label{fig:humaneval_qfs}
\end{figure}
\subsection{Comparing to Pre-Finetuning}
Transferring information from related tasks is another approach to adapt generic models to specific tasks \citep{aghajanyan-etal-2021-muppet}.
We show in \autoref{tab:orig_results} that \socratic{} pretraining outperforms even the best pre-finetuned BART SegEnc model, which uses additional supervision from the WikiSum dataset \citep{liu2018generating}.
This transfer dataset was selected from a wide range of relevant summarization datasets tested by \citet{vig-etal-2022-exploring}. 
Crucially, we note that transfer learning, like pre-finetuning, is orthogonal to our line of work which operates on the pretraining side. We believe that \socratic{} can therefore be used in combination with pre-finetuning to further boost performance.
\begin{figure}
    \centering
    \includegraphics[width=0.75\columnwidth, trim= {0 0.5cm 0 0}, clip]{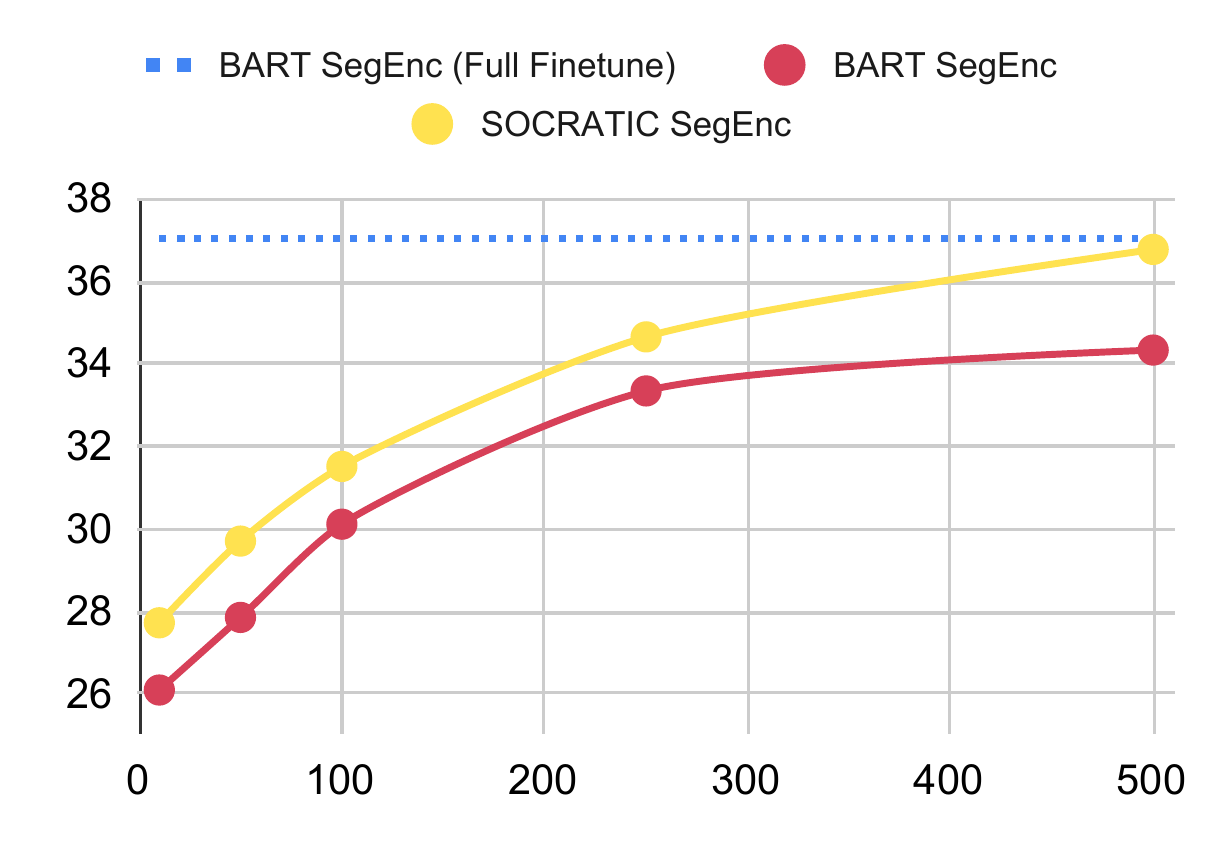}
    \caption{Few-shot performance on QMSum test set.}
    \label{fig:few_shot}
\end{figure} 

\subsection{General vs. Specific Summaries}
Both QMSum and Squality datasets contain a substantial portion of general summaries (12.5-20\%) that aim to summarize the entire document in addition to those answering more specific queries. We find that our approach improves in both cases (+0.98 and +0.28 ROUGE-1 on QMSum in general and specific queries respectively). This shows that \socratic{} pretraining improves models intended to perform a combination of general-purpose and query-focused summarization. In addition, with users increasingly interacting with language models through prompts to perform different tasks, the query-focused datasets we evaluate on become realistic testbeds for NLP systems that aim to perform well across tasks.

\subsection{Few-Shot Finetuning}
To show that \socratic{} pretraining alleviates the need for labeled downstream task data, we study the few-shot learning performance of \socratic{} and BART SegEnc models. We perform one finetuning run for each model on each subset of the task data.
In \autoref{fig:few_shot}, we show that with half the QMSum examples, \socratic{} SegEnc achieves the same performance as finetuning BART SegEnc on all of QMSum. 
We believe that bringing \socratic{} pretraining closer to the downstream task of query-focused summarization lets the models learn from fewer downstream task examples.

\section{Finegrained Planning Results}
In this section, we evaluate the effect of \socratic{} pretraining on the adherence to user-provided finegrained control sequences. 
In these experiments, the same \socratic{} pretrained model is finetuned on task-specific data with various control strategies.


\subsection{Going Beyond High-Level Questions}
\label{sec:control_strategies}
The queries found in QMSum and \squality{} are only one format to encode user intent.
Previous research explored other control strategies like keywords \cite{he2020ctrlsum}, entity chains \citep{narayan-etal-2021-planning}, or factoid question-answer pairs \citep{narayan2022conditional}. As seen in \autoref{fig:finegrained-control}, these strategies offer a more finegrained level of control over the summaries as they operate at the sentence level. 
Reference control sequences are not available for QMSum and \squality{} so we \emph{generate them automatically} from reference summaries. 
In the summarization literature, 
such control sequences are often modeled as intermediate plans generated before the summaries~\citep{narayan2022conditional, he2020ctrlsum}. 
In these cases, given the input $X$, the model first generates the detailed plan for the summary $B$ from $P(B|X)$, then generates the summary $Y$ conditioning on the plan and the input $x$ from $P(Y|B,X)$.
Even if the plan $B$ is initially generated by the model, a user can control the summary by altering the plan. 
In practice, we experiment with three different planning strategies.

\begin{itemize}
    \item \emph{Content questions}. For each sentence in the reference summary, we generate a question using the MixQG system while giving the full summary as context. 
    These are similar to the questions that we use in our \socratic{} pretraining. The sentence-level questions are then concatenated into a single plan for the summary. To our knowledge, we are the first to propose using content questions as finegrained plans for summaries.
    \item \emph{QA blueprint}. We reimplement the recently proposed text plan in the form of a sequence of question-answer (QA) pairs \citep{narayan2022conditional}. 
    First, all noun phrase answers are extracted from the reference. Then, a QG system generates questions answered by each noun phrase. The QA pairs are then filtered using round-trip consistency, rheme, and coverage criteria. The final plan consists of the concatenation of the remaining QA pairs.
    \item \emph{Keywords}. We use keywords extracted from each sentence of the reference summary. We take the noun-phrase answers from the QA blueprint as keywords and concatenate them with sentence separators into a plan. 
\end{itemize}

\label{sec:finegrained-planning}
\begin{figure}[t!]
  \center{\scriptsize
  \setlength\tabcolsep{0.1cm}
    \begin{tabular}{p{7.5cm}}
    \toprule 
    \textbf{Original Text:} In a group discussion about a philosophical concept, Sarah used the Socratic method by asking and answering questions to stimulate critical thinking and clarify underlying assumptions. The method helped her and her classmates achieve a deeper understanding of the concept and address disagreements. Sarah looked forward to continuing to use it in her studies. \\
    \midrule
    \textbf{Content Questions (Ours):} How did Sarah use the Socratic method? What were the benefits of the Socratic method? What did Sarah think of the method? \\
    \midrule
    \textbf{Keywords:} Group discussion | Sarah | Socratic method | questions | thinking | assumptions || method | classmates | understanding | disagreement ||  studies \\
    \midrule
    \textbf{Blueprint QA:} What type of discussion did Sarah have about a philosophical concept? Group discussion | Who used the Socratic method? Sarah | What method did Sarah use to stimulate critical thinking? Socratic method | What did Sarah ask in the Socratic method? questions | What did Sarah clarify in the Socratic method? assumptions ...
    \\
    \bottomrule
    \end{tabular}     
  }
  \caption{Comparison of finegrained control strategies.}
  \label{fig:finegrained-control}
\end{figure}

\begin{table*}[ht]
    \centering\scriptsize
    \resizebox{\textwidth}{!}{
    \begin{tabular}{ll@{\hskip 0.3in}cccc@{\hskip 0.4in}cccc}
    \toprule
    &  &  \multicolumn{4}{c}{Summary} & \multicolumn{4}{c}{Control Plan} \\
    Control Strategy & Model & Rouge1 & Rouge2 & RougeL & BS-R & Rouge1 & Rouge2 & RougeL & Leven. Edit \\
    \midrule
    \multirow{3}{*}{ Content Questions } & BART-Large SegEnc & 35.3 & 11.6 & 30.7 & 86.95 & 42.3 & \textbf{23.4} & 41.6 & 0.77 \\
    & + PEGASUS Pret. & 35.4 & 11.8 & 30.9 & 87.03 & 41.7 & 22.9 & 41.0  & 0.74 \\
    & + \socratic{} Pret. & \textbf{36.0} & \textbf{12.1} & \textbf{31.5} & \textbf{87.15} & \textbf{42.4} & 23.2 & \textbf{41.7} & 0.77\\
    \midrule
    \multirow{2}{*}{ Blueprint QA } & BART-Large SegEnc & 33.5 & 9.3 & 29.4 & 86.62 & 40.2 & 15.7 & 39.2 & 0.85\\
     & + \socratic{} Pret. & \textbf{35.4} & \textbf{10.0} & \textbf{30.6} & \textbf{86.89} & \textbf{40.7} & \textbf{15.9} & \textbf{39.6} & 0.85 \\
    \midrule
    \multirow{2}{*}{ Keywords } & BART-Large SegEnc & 36.2 & 12.8 & 31.4 & \textbf{87.01} & 24.1 & 9.2 & 21.3 &  0.88\\
    & + \socratic{} Pret. & \textbf{36.9} & \textbf{13.2} & \textbf{32.1} & \textbf{87.01} & \textbf{25.0} & \textbf{10.0} & \textbf{22.1} & 0.88 \\
    \bottomrule
    \end{tabular}}
    \caption{Results on different control strategies on QMSum (results averaged over five random seeds).}
    \label{tab:qmsum_q_s}
\end{table*}
\subsection{Comparing Control Strategies}
In \autoref{tab:qmsum_q_s}, we report evaluation metrics for both the model-generated summaries and plans. 
\par 
We find that with all three control strategies, \socratic{} pretraining provides a consistent improvement over the vanilla BART model and the PEGASUS pretraining on both the generated finegrained plan and summary. 
On the planning side, there is a small but consistent improvement, up to +0.9 Rouge-1 with keyword chain control, indicating that the model has improved planning abilities.
On the summarization side, we find a more significant improvement with up to +1.9 Rouge-1 with blueprint QA control. We attribute this to a combination of improved planning and execution ability of the model from \socratic{} pretraining.

With respect to control strategy performance, we find that our content questions obtain the highest Rouge scores (42.4 Rouge-1), outperforming keyword chains with only 25.0 Rouge-1. 
Despite the keyword plan having low overlap with the reference, it results in good summarization performance, so it is unclear whether the model using keyword chains learns the right correspondence between plan and summary. 
Moreover, the generated keyword chain would need heavier editing to obtain the reference plan compared to the content question plan (0.88 Levenstein distance compared to 0.77), making them less useful in practice. 

Previous work has focused on keyword controls \citep{he2020ctrlsum} and fact-oriented questions for text generation \citep{narayan2022conditional}, but there are inherent limitations with these approaches, which we discuss in detail in \ref{sec:compare-control}.

\begin{table}[t!]
    \centering
    \resizebox{\columnwidth}{!}{
    \begin{tabular}{llcccc}
    \toprule
    Oracle Strategy& Model & R-1 & R-2 & R-L & BS-R \\
    \midrule
    \multirow{2}{*}{ Content Questions } & BART-Large SegEnc & 43.7 & 18.0 & 39.0 & 88.32 \\
    & + \socratic{} Pret. & \textbf{46.8} & \textbf{20.3} & \textbf{41.7} & \textbf{88.92} \\
    \midrule
    \multirow{2}{*}{ Blueprint QA } & BART-Large SegEnc & 52.9 & 24.1 & 46.8 & 89.63 \\
    & + \socratic{} Pret. & \textbf{56.3} & \textbf{26.6} & \textbf{49.3} &  \textbf{90.03} \\
    \midrule
    \multirow{2}{*}{ Keywords } & BART-Large SegEnc & 45.7 & 20.2 & 40.5 & 88.73 \\
   & + \socratic{} Pret. & \textbf{47.5} & \textbf{21.9} & \textbf{42.5} & \textbf{89.18} \\
    \bottomrule
    \end{tabular}
    }
    \caption{Performance on the QMSum dataset with various oracle finegrained control strategies. 
    }
    \label{tab:qmsum_q_s_oracle}
\end{table}

\begin{figure}[t!]
    \centering
    \includegraphics[width=1.0\columnwidth, trim= {0 0.8cm 0 0}, clip]{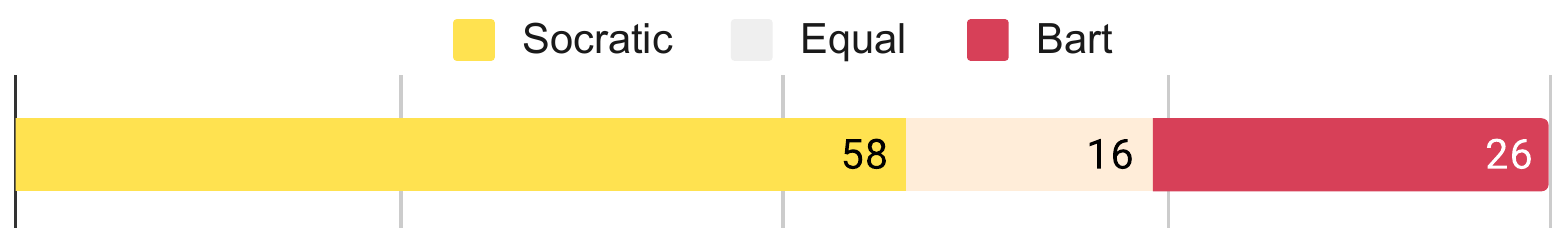}
    \caption{Annotators' finegrained planning preferences.}
    \label{fig:humaneval_qfs_finegrained}
\end{figure}
\subsection{Oracle Questions}
Ideally, users can tailor generated summaries with an intervention limited to editing the generated plans.
However, this requires strong adherence of generations to the finegrained plans, which we test here with oracle plans.
Instead of generating both plan and summary, the system is given the oracle plans automatically extracted from the reference summaries (see \ref{sec:control_strategies}).
In \autoref{tab:qmsum_q_s_oracle}, we observe a large improvement of +3.1 Rouge-1 over the BART SegEnc baseline. Human annotators confirm that \socratic{} SegEnc follows oracle finegrained plans better or similarly to the baseline in 74\% of instances, shown in \autoref{fig:humaneval_qfs_finegrained} and described further in \ref{sec:human-evaluation}.
This confirms our hypothesis that \socratic{} pretraining helps ground the generations to user-provided queries. We attribute these gains to using the \askanswer{} mode, 
which introduces structure in the pretraining data by using as target text a question plan followed by its pseudo-summary answer.
We hypothesize that this structure in pretraining is what helps the model adhere to the planning step more effectively regardless of the control strategy.

\section{Conclusion}
In this work, we introduce \socratic{} pretraining, a question-driven, unsupervised pretraining objective to adapt generic language models to the task of controllable summarization. 
\socratic{} pretraining trains the model to generate relevant questions in a given context and then to answer them. 
Our experiments demonstrate the generality of our approach both on query-focused summarization and finegrained controllable summarization.
We show that \socratic{} pretraining outperforms other pretraining and prefinetuning objectives, that it cuts downstream task data requirements in half, and that it works across control strategies and domains.

\section{Limitations}
\paragraph{Downstream Tasks}
In this work, we focused on long-document summarization as we believe it is the task where controllable summarization is most needed. Future work could investigate the effect of \socratic{} pretraining on other downstream applications beyond those studied here. To handle long document input we could not use the BART model with \socratic{} pretraining adaptation directly. Instead, we applied the SegEnc architecture on top of BART. 
This adaptation of the pretrained model may have dampened some of the few-shot performance of \socratic{} pretraining. 
We thus believe that tasks with shorter input documents for which the SegEnc architecture is not necessary would see even greater benefits in the low-resource setting.

\paragraph{Base Model}
Throughout this work, we restricted our analysis to one model architecture the SegEnc architecture with the BART base model. Previous work extensively studied the impact of different architectures for long-document query-focused summarization \citep{vig-etal-2022-exploring}. These primarily differ in how they model long documents. The authors found SegEnc, a simple sliding window adaptation of BART, to perform best on QMSum. While the results presented here are specific to SegEnc and BART, our approach is agnostic to the underlying model architecture and is orthogonal to long-document modeling. We leave it to future work to investigate the effect \socratic{} pretraining has on other architectures.

\paragraph{Evaluation Metrics} As discussed in prior work \citep{fabbri-etal-2021-summeval, pagnoni-etal-2021-understanding, gehrmann-etal-2021-gem}, there are limitations with the current automated evaluation metrics which do not strongly correlate with human judgments. Our results from these metrics should therefore be interpreted with caution and in combination with the human evaluation we performed to support them. One area in which automated metrics have been reported to perform poorly is factuality. Moreover, current factuality metrics have been designed and tested in the news domain and their performance in the out-of-domain setting (long documents and dialog data) was not systematically evaluated and is hard to interpret \citep{agarwal-etal-2022-creativesumm}. In this work, we therefore choose not to report any factuality metric results.

\paragraph{QG Efficiency} We did not optimize the efficiency of the QG component of \socratic{} pretraining and, consequently, it is computationally expensive. Currently, given equal amounts of resources for QG and pretraining, it takes us about the same time to perform the QG phase and pretraining phase on the same amount of data. We note, however, that in low-resource scenarios, the additional compute can lead to significant benefits, as shown in our results. In addition, we did not experiment with efficient sampling strategies, and believe that improving the efficiency of the QG model inference, for example through model distillation \citep{hinton2015distilling}, could lead to significant efficiency gains. 

\paragraph{Dataset Biases}
The datasets for pretraining and finetuning used in this work are in English and thus mainly represent the culture of the English-speaking populace. 
Political or gender biases may also exist in the dataset, and models trained on these datasets may propagate these biases. 
Additionally, the pretrained BART model carries biases from the data it was pretrained on. 
We did not stress test these models for biases and request that the users be aware of these potential issues in applying the models presented. 
\paragraph{Misuse Potential and Failure Mode}
When properly used, the summarization models described in this paper can be time-saving.
However, the current model outputs may be factually inconsistent with the input documents, and in such a case could contribute to misinformation on the internet. 
This issue is present among all current abstractive summarization models and is an area of active research.


\bibliography{anthology,custom}
\bibliographystyle{acl_natbib}

\appendix
\newpage
\section{Appendix}
\label{sec:appendix}

\subsection{Dataset Information}
We use the QMSum and \squality{} datasets according to their intended research purposes.
\label{sec:dataset-info}
\begin{table}[ht]
    \centering
    \resizebox{\columnwidth}{!}{
    \begin{tabular}{llrrr}
    \toprule
     Dataset & Domain & \# Ex. & Doc. Len & Sum. Len \\
     \midrule
     CNN/DM & news & 311K & 804 & 60 \\
     XSum & news & 226K & 438 & 24\\
     \midrule
     QMSum & meetings & 1,808 & 9,067 & 70\\
     \squality{} & stories & 625 & 5,200 & 237 \\
    \bottomrule
    \end{tabular}}
    \caption{Statistics of general summarization vs. QFS datasets, length in words \citep{wang2022squality}.}
    \label{tab:downstream_tasks}
\end{table}

\subsection{Training Details}
\label{sec:training-details}
We describe here the training details for \socratic{} pretraining as well as downstream task finetuning. Our experiments rely on the Huggingface Transformers library \citep{wolf-etal-2020-transformers}. Our code includes sample pretraining and finetuning scripts to facilitate the reproduction of our results. We use 8 Nvidia A100 GPUs to run the experiments described in this paper. We will release our code under BSD 3-Clause license.

\subsubsection{Pretraining}
\paragraph{Data Preprocessing} In the Books3 corpus, documents are longer than the desired input and target texts, we therefore segment the documents to obtain roughly the desired lengths. In the UnDial dataset, the opposite is true and therefore we concatenate dialogues to obtain the desired lengths. Following this segmentation or concatenation, we mask the input text and construct the target as described in \autoref{sec:socratic-pretraining} depending on the desired mode. We then truncate the input and target texts to 256 and 512 tokens respectively.

\paragraph{Special Tokens} We introduce mode tokens and a new separator tokens to the tokenizer of the BART-large model before the pretraining adaptation step.

\paragraph{Training Hyperparameters}
We train the BART-large model for 100k steps with batch size 512, checkpointing every 10k steps. For the ablations, we use batch size of 64 and the same number of steps. In all our experiments, we use AdamW optimizer with 5k warmup steps, learning rate 3e-5, weight decay of 0.01, max grad norm of 0.1, and bfloat16. Our choice of hyperparameters is based on best practices from previous work performing pretraining adaptations of BART-large \citep{xiao-etal-2022-primera, wan-bansal-2022-factpegasus}. We also performed grid-search on the learning rate on the small-scale pretraining dataset testing the values \{3e-6, 3e-5, 1e-4\} but finding the initial value to perform best. We use the same hyperparameters on all three pretraining corpora in our ablations. 

\paragraph{Checkpoint validation} We evaluate the checkpoints on the validation dataset of the target downstream tasks and pick the best performing checkpoint.

\subsubsection{Finetuning}
\paragraph{SegEnc Implementation} We use the SegEnc implementation from the original authors. Instead of using vanilla BART-large to initialize the SegEnc model, we use one of our pretrainined models. 

\paragraph{Finetuning Hyperparameters} We use the same hyperparameters for both QMSum and \squality{} datasets and for QFS and finegrained planning experiments. We train the SegEnc model for 10 epochs with batch size 1 and bfloat16. We use the AdamW optimizer with learning rate 5e-6. We tested the following learning rate values \{5e-7, 5e-6, 5e-5, 5e-4\}. We use beam search decoding with beam size of 4. Our hyperparameters follow the best performing hyperparameters found by the original authors of the SegEnc model \citep{vig-etal-2022-exploring}. Annotations will be made available ensuring the identity of the workers remains anonymous. We will only report the answers to the questions for each example and anonymize the worker ID. 

\paragraph{Mode} While the \socratic{} pretraining consists of both \emph{Reconstruct} and \askanswer{} modes, we found that the latter performed best on the downstream tasks.

\subsection{Automated Evaluation Details}
We perform an automated evaluation using Rouge and BERTScore metrics following best practices from previous work. Specifically, we use the evaluation setup from \citet{vig-etal-2022-exploring} for QMSum and the evaluation setup from \citet{wang2022squality} for \squality{}. More details and the relevant scripts can be found the in the supporting code supporting their papers. We also provide scripts to reproduce our evaluation. For BERTScore, we report recall following recommendations from \citet{liu2022revisiting}.

\subsection{Human Evaluation Details}
\label{sec:human-evaluation}
We perform a human evaluation study to confirm that variations between models are perceptible and meaningful to human users.
The study separately assesses the QFS and the finegrained planning models finetuned on the QMSum dataset. In both cases, we use 100 of the 281 examples from the QMSum test set, and three independent annotators from the Amazon Mechanical Turk platform. We restrict the study to the specific questions of the QMSum dataset as these also provide relevant text spans in the original dialogue.

We measure inter-annotator agreement with Fleiss Kappa $\kappa$ \citep{fleiss1971measuring} and obtain fair to moderate agreement in our tasks. Other studies that also rely on untrained crowd-sourced workers report similar, or sometimes even lower, agreement \citep{goyal2022news}.

\begin{figure*}
    \centering
    \includegraphics[width=\textwidth, trim= {8cm 30cm 8cm 0}, clip]{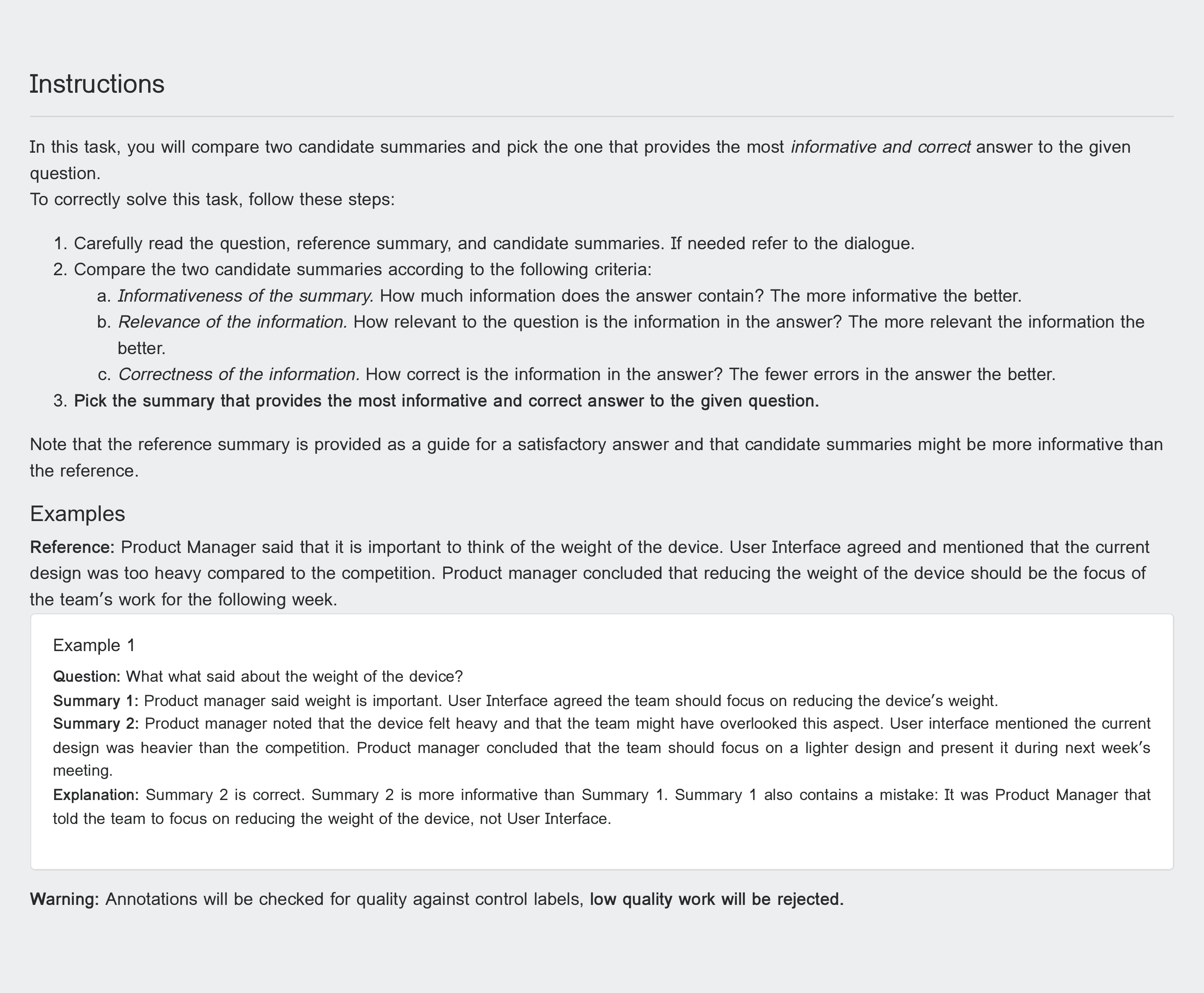}
    \caption{QFS human annotation instructions.}
    \label{fig:qfs_platform}
\end{figure*}

\paragraph{QFS Task}
In this task, we compare the SegEnc model with \socratic{} pretraining to Pegasus and BART pretraining.
We ask annotators to select the best answer to the given query  between two candidate summaries or mark if they are equally good. We provide both the reference summary and the relevant text span as supporting information. Annotator agreement on this task is $\kappa=0.33$. The results are summarized in \autoref{fig:humaneval_qfs} and the annotation instructions can be found in \autoref{fig:qfs_platform}.

\begin{figure*}
    \centering
    \includegraphics[width=\textwidth, trim= {12cm 50cm 12cm 0}, clip]{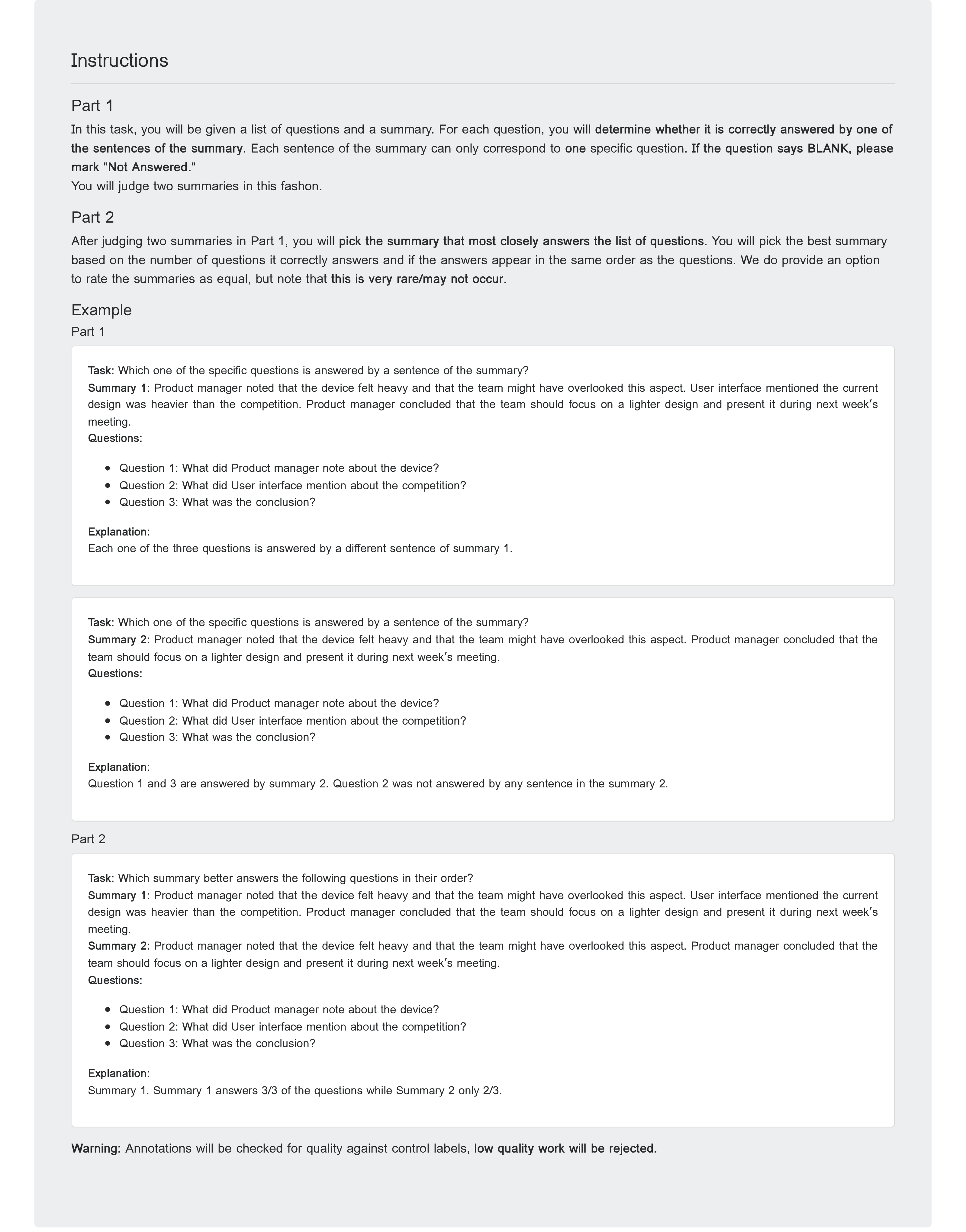}
    \caption{Finegrained planning human annotation instructions.}
    \label{fig:Finegrained_qfs_platform}
\end{figure*}

\paragraph{Finegrained Planning Task}
In this task, we compare the \socratic{} SegEnc model to the baseline BART SegEnc model in terms of their adherence to a finegrained plan. Both models are finetuned to the finegrained planning task on QMSum with the \emph{content question} control strategy. Here we test how well they follow oracle plans automatically generated from the reference summary. The task is structured in two parts. First, for each question of the oracle plan, we ask annotators whether a sentence of the summary answers the question. We repeat for both \socratic{} and BART summaries. On this task, we obtain moderate agreement of $\kappa=0.49$. Next, we ask the annotators to select the best summary between the two candidates in terms of how closely it follows the plan. For the second task, the agreement is $\kappa=0.34$.
The results are summarized in \autoref{fig:humaneval_qfs_finegrained} and the annotation instructions can be found in \autoref{fig:Finegrained_qfs_platform}.

\paragraph{Worker Selection and Considerations}
An ethics review board did not review this particular protocol, but we followed prior protocols and internally-approved guidelines, such as carefully calibrating the time/HIT to ensure a pay-rate of \$12/hour and letting workers know that their annotations will be used as part of a research project to evaluate the performance of summarization systems.

We selected workers according to the following criteria: HIT approval rate greater than or equal to 98\%, number of HITs approved greater than or equal to 10000, and located in either the United Kingdom or the United States. The workers also passed a qualification test for a related summarization task from a prior project, ensuring that the annotators were familiar with the task of judging model-generated summaries.

\subsection{Comparing Control Strategies}
\begin{table*}[ht!]
    \centering
    \resizebox{\textwidth}{!}{
    \begin{tabular}{lcccc}
         \toprule
         \multirow{2}{*}{Control Type} & Length & Lexical Overlap With Summ. & \multicolumn{2}{c}{Lexical Overlap Across Queries}  \\
         & \% of summ. len & Rouge 1 & Avg. Overlap & Max. Overlap \\
         \midrule
         Keywords & 25\% & 37.9 & 43\% & 100\% \\
         Bleuprint QA & 149\% & 65.9  & 22\% & 44\% \\
         Content Questions (ours) & 48\% & 38.1 & 36\% & 67\% \\
         \bottomrule
    \end{tabular}
    }
    \caption{Properties of finegrained control strategies for the QMSum dataset. We measure lexical overall between the control sequence and the reference summary. We also calculate the average and maximum lexical overlap of two control sequences from the same QMSum  document but answering two different high-level queries.}
    \label{tab:guidance_analysis}
\end{table*}
\label{sec:compare-control}
Using content questions for QG augmentation in \socratic{} pretraining improves performance across control strategies, including on non-question-based finegrained controls like keyword chains (see \autoref{tab:qmsum_q_s}). While most previous work has focused on keyword controls \citep{he2020ctrlsum} and fact-oriented questions for text generation \citep{narayan2022conditional}, there are inherent limitations with these approaches. 
We identify important qualitative properties of queries for controllable generation below that informed our choice of content questions for \socratic{} pretraining.

\paragraph{Natural}
To facilitate the use of controllable summarization, one overarching objective is to make the user interaction with the system as natural as possible. When evaluating how ``natural'' a query strategy is, we consider whether such a strategy is used by humans when they interact with one another. According to this perspective, using keywords is an unnatural query strategy. Users generally express themselves through natural language, and when inquiring about information, they use questions. Our query systems in controllable summarization should strive to reflect this and support natural queries from the users.

\paragraph{Unambiguous}
To ensure that summaries contain the intended information, it is necessary that queries refer with minimal ambiguity to the information of interest in the document. When dealing with long documents, where the same entities occur repeatedly, keywords often imprecisely describe the intended query.  But it is precisely with such long documents that query-focused summarization is particularly useful. In \autoref{tab:guidance_analysis}, we show that different keyword queries about the same document have a lexical overlap of 46\% of words on average and 100\% in the worst-case scenario in QMSum. In comparison, content questions have a word overlap of 36\% on average and no more than 67\%. When formulating queries in natural language, they more richly encode the entities and their relations making them less ambiguous.

\paragraph{Concise}
Fact-oriented question-answer pairs (blueprint QA) \citep{narayan2022conditional} tend to be less ambiguous than keywords (with the least lexical overlap across the three query strategies) but often end up requiring more text than the summary itself. On average, blueprint QA uses 50\% more words than the summary (see \autoref{tab:guidance_analysis}). This makes this query strategy impractical for controllable summarization where the concision of the query is a desirable property.


\end{document}